\documentclass{article} 

\usepackage{longtable}
\usepackage{ragged2e} 
\usepackage{booktabs}
\usepackage[load-configurations=version-1]{siunitx} 

\title{Discovering Association Rules in High-Dimensional Small Tabular Data\thanks{Published version available at \url{https://ceur-ws.org/Vol-4125/paper_26.pdf}}}

\usepackage{glossaries}
\usepackage{multirow}
\usepackage{makecell}
\usepackage{float}
\usepackage{wrapfig}
\usepackage{siunitx}
\usepackage[justification=centering]{caption}
\usepackage{enumitem}
\usepackage{acronym}
\usepackage{natbib}
\usepackage{arxiv}
\usepackage[ruled]{algorithm2e}
\usepackage{amssymb}

\usepackage{tabularx}
\usepackage{multirow}
\usepackage{multicol}
\usepackage[utf8]{inputenc} 
\usepackage[T1]{fontenc}    
\usepackage{hyperref}       
\usepackage{url}            
\usepackage{booktabs}       
\usepackage{amsfonts}       
\usepackage{nicefrac}       
\usepackage{microtype}      
\usepackage{amsmath}
\usepackage{cleveref}       
\usepackage{lipsum}         
\usepackage{graphicx}
\usepackage{natbib}
\usepackage{doi}
\usepackage{authblk}

\usepackage[T1]{fontenc}
\usepackage{graphicx}
\usepackage{booktabs}
\usepackage[misc]{ifsym}

\usepackage{graphicx}
\usepackage{blindtext}
\usepackage{glossaries}
\usepackage{mathtools}
\usepackage{wrapfig}
\usepackage{wrapfig}
\usepackage[noend]{algpseudocode}
\usepackage{multirow}
\usepackage{subfig}
\usepackage[table, svgnames, dvipsnames]{xcolor}
\usepackage{makecell, cellspace}
\usepackage{fancyhdr}

\usepackage[font=small,labelfont=bf,
   justification=justified,
   format=plain]{caption}

\acrodef{DL}{Deep Learning}
\acrodef{ML}{Machine Learning}
\acrodef{CNF}{Conjunctive Normal Form}
\acrodef{ARM}{Association Rule Mining}
\acrodef{NARM}{Numerical Association Rule Mining}
\acrodef{ANN}{Artificial Neural Network}
\acrodef{IoT}{Internet of Things}

\newbox{\orcid}\sbox{\orcid}{\includegraphics[scale=0.06]{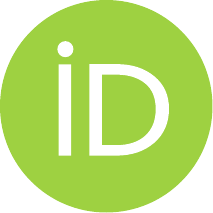}}

\fancypagestyle{firstpage}{
    \fancyhf{}
    
    \fancyhead[C]{\footnotesize This paper was accepted at ECAI 2025 Workshop: 1st International Workshop on Advanced Neuro-Symbolic Applications (ANSyA)}
}

\author[1]{%
    \href{https://orcid.org/0000-0003-2710-7951}{\usebox{\orcid}\hspace{1mm}Erkan Karabulut}\thanks{Corresponding author: \texttt{e.karabulut@uva.nl}}%
}
\author[2]{%
    \href{https://orcid.org/0000-0002-5357-3705}{\usebox{\orcid}\hspace{1mm}Daniel Daza}%
}
\author[1]{%
    \href{https://orcid.org/0000-0003-0183-6910}{\usebox{\orcid}\hspace{1mm}Paul Groth}%
}
\author[1]{%
    \href{https://orcid.org/0000-0001-7054-3770}{\usebox{\orcid}\hspace{1mm}Victoria Degeler}%
}

\affil[1]{University of Amsterdam, The Netherlands}
\affil[2]{Amsterdam University Medical Center, The Netherlands}

\begin{document}

\maketitle
\thispagestyle{firstpage}

\begin{abstract}
Association Rule Mining (ARM) aims to discover patterns between features in datasets in the form of propositional rules, supporting both knowledge discovery and interpretable machine learning in high-stakes decision-making. However, in high-dimensional settings, rule explosion and computational overhead render popular algorithmic approaches impractical without effective search space reduction—challenges that propagate to downstream tasks. Neurosymbolic methods, such as Aerial+, have recently been proposed to address the rule explosion in ARM. While they tackle the high-dimensionality of the data, they also inherit limitations of neural networks, particularly reduced performance in low-data regimes.

This paper makes three key contributions to association rule discovery in high-dimensional tabular data. First, we empirically show that Aerial+ scales one to two orders of magnitude better than state-of-the-art algorithmic and neurosymbolic baselines across five real-world datasets. Second, we introduce the novel problem of ARM in high-dimensional, low data settings, such as gene expression data from the biomedicine domain with \textasciitilde18K features and \textasciitilde50 samples. Third, we propose two fine-tuning approaches to Aerial+ using tabular foundation models. Our proposed approaches are shown to significantly improve rule quality on five real-world datasets, demonstrating their effectiveness in low-data, high-dimensional scenarios.
\end{abstract}

\section{Introduction}

\ac{ARM} is the task of discovering patterns among the features of a dataset in the form of logical implications~\citep{agrawal1994fast}, also known as if-then rules. ARM has been applied in a myriad of domains for knowledge discovery~\citep{luna2019ARMsurvey} as well as for high-stakes decision-making as part of interpretable machine learning models~\citep{rudinnature,corels}. High-dimensional datasets, e.g., with thousands of columns, often lead to rule explosion and prolonged execution times~\citep{moens2013frequent}. Common solutions to rule explosion in ARM include constraining data features (i.e., ARM with item constraints~\citep{srikant1997mining,baralis2012generalized,yin2022constraint}), mining top-k high-quality rules~\citep {fournier2012mining,nguyen2018etarm}, and closed itemset mining~\citep{zaki2002charm}. However, these methods mainly focus on reducing the search space for knowledge discovery, rather than directly addressing the computational burden. 

Neurosymbolic methods for ARM, such as Aerial+~\citep{aerial_plus}, have been recently proposed to address the rule explosion problem on tabular data. Despite its effectiveness in addressing the rule explosion problem in generic tabular data, Aerial+ has not yet been evaluated on high \textit{d}-dimensional datasets for scalability. Moreover, neurosymbolic methods for ARM also inherit the limitations of neural networks, such as reduced performance in low-data (n) regimes~\citep{liu2017deep}.

\begin{table}[t]
    \centering
    \caption{\textbf{Sample $d \gg n$ dataset and association rules.} Gene expression datasets in tabular form often consist of 10K+ columns and a limited number of rows. This is a sample gene expression level data from \citep{gao2015high}, partially pre-processed by \citep{ruiz2023high} and put in discrete form after applying z-score binning. Listed association rules are learned using Aerial+~\citep{aerial_plus} with item constraints on low and high values.}
    \vspace{5pt}
    \begin{tabular}{lcccccc}
        \toprule
        \textbf{Sample / Rule} & \textbf{Gene\_1} & \textbf{Gene\_2} & \textbf{Gene\_3} & $\cdots$ & \textbf{Gene\_18107} & \textbf{Gene\_18107} \\
        \midrule
        Sample\_1 & normal & normal & normal & $\cdots$ & normal & normal \\
        Sample\_2 & normal & normal & high & $\cdots$ & normal & high \\
        Sample\_3 & normal & normal & normal & $\cdots$ & normal & low \\
        \multicolumn{7}{c}{$\cdots$} \\
        \midrule
        \midrule
        \textbf{Rule\_1} & \multicolumn{6}{c}{Gene2 (high) $\wedge$ Gene29 (high) $\rightarrow$ Gene14 (low)} \\
        \textbf{Rule\_2} & \multicolumn{6}{c}{Gene3 (high) $\wedge$ Gene45 (high) $\rightarrow$ Gene84 (high)} \\
        \bottomrule
    \end{tabular}
    \label{tab:motivation}
\end{table}

As is the case for several data-driven methods, Aerial+ relies on statistical patterns present in the dataset. In small datasets, such patterns may be hard to extract, which in turn may lead to reduced predictive performance, and in the case of Aerial+, to rules that do not accurately capture the true underlying patterns. Recent works on models for tabular data have addressed this issue by introducing \emph{foundation models}~\citep{tabpfn,qu2025tabicl,iida2021tabbie,yin2020tabert,su2024tablegpt2}, which are pre-trained on large datasets and transferred to small datasets without additional training, thereby providing strong inductive biases and generalizable representations that compensate for the limited data instances.

In this paper, we make three key contributions to ARM research on categorical tabular data. First, we evaluate the scalability of both commonly used algorithmic ARM approaches as well as the recent Neurosymbolic methods on high \textit{d}-dimensional datasets. Second, to the best of our knowledge, we introduce the problem of ARM on $d \gg n$ datasets for the first time, which are common in the biomedicine domain, such as gene expression datasets~\citep{gao2015high} (see Table \ref{tab:motivation}), and evaluate the recent neurosymbolic methods on such datasets in terms of rule quality. Third, we propose two fine-tuning methods for neurosymbolic ARM methods that rely on tabular foundation models for addressing the low-data regime.

Our empirical results show that: i) Aerial+ scales one to two orders of magnitude faster on high-dimensional datasets compared to state-of-the-art ARM methods (Section \ref{sec:arm-on-high-dim-data}), ii) neurosymbolic methods need longer training to find high-quality association rules on $d \gg n$ datasets (Section \ref{sec:arm-in-low-data}), iii) our two proposed fine-tuning methods allow Aerial+ to learn significantly higher quality rules in small datasets (Section \ref{sec:arm-in-low-data}). The results indicate that neurosymbolic methods, especially when supported with tabular foundation models, can enable scalable and high-quality knowledge discovery in high-dimensional tabular data with few instances (Section \ref{sec:discussion}).  

\section{Related Work} \label{sec:related-work}

This section presents a formal definition of ARM on categorical tabular data, the problem of high-dimensional data with few instances, neurosymbolic ARM methods, and tabular foundation models.

\textbf{Association rule mining.} Following the original definition of ARM in~\citep{agrawal1994fast}, let $I=\{i_1, i_2, ..., i_m\}$ be a set of $m$ items, and let $D = \{t_1, t_2, ..., t_n\}$ be a set of $n$ transactions where $\forall t \in D, t \subseteq I$ meaning each transaction $t$ consists of a set of items in $I$. An association rule is of the form $X \rightarrow Y$, where $X,Y \subseteq I$, is a first-order Horn clause with at most one positive literal, $|Y| = 1$ and $|X| \ge 1$, in its \ac{CNF} ($\lnot X \lor Y$), and $X \cap Y = \varnothing$. Note that $p \rightarrow q \wedge r$ can be rewritten as $p \rightarrow q$ and $p \rightarrow r$, (i.e., $p, q, r \in I$). $X$ is often referred to as the \textit{antecedent} while $Y$ is the \textit{consequent} side of the association rule. Example association rules are given in Table \ref{tab:motivation}. A rule $X \rightarrow Y$ is said to have \textit{support} percentage $s$ if s\% of $t \in D$ contain $X \cup Y$, while the \textit{confidence} of a rule is defined as $\frac{support(X \rightarrow Y)}{support(X)}$. ARM has initially been defined as the problem of finding rules that have higher minimum support and confidence values than a given user-defined threshold. The state-of-the-art in ARM literature has a plethora of sub-problems and solutions which can be found in \citep{luna2019ARMsurvey,kaushik2023numerical}. Categorical tabular data is often converted to a set of transactions via one-hot encoding, where each encoded value represents the presence (1) or absence (0) of a column-value pair, corresponding to items in $I$, and each row corresponds to a transaction in $D$.

\begin{figure*}[t]
    \centering
    \includegraphics[width=\linewidth]{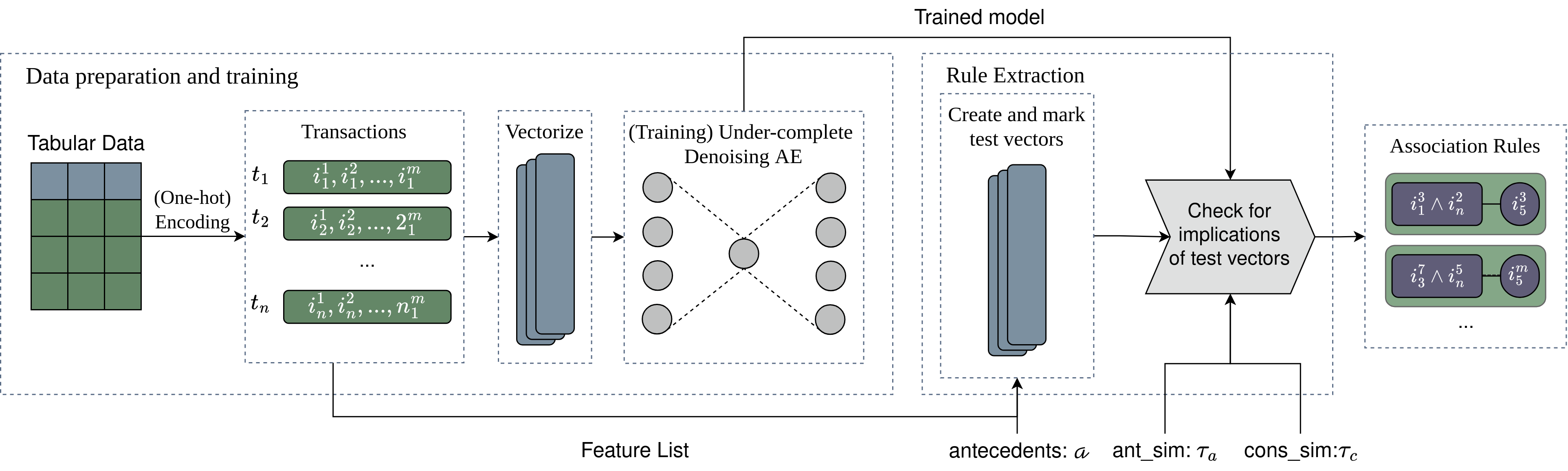}
    \caption{Aerial+~\citep{aerial_plus} ARM pipeline consists of: i) converting given categorical tabular data into transactions by one-hot encoding, ii) vectorizing the one-hot encoded data, iii) training an under-complete denoising Autoencoder with a reconstruction loss and to output probability distributions per column, iv) and extracts association rules by exploiting the reconstruction ability of autoencoders, based given probabilistic antecedent and consequent similarity thresholds.}
    \label{fig:aerial-pipeline}
\end{figure*}

\textbf{ARM for high-dimensional small data.} Having high-dimensional data with a limited number of samples is common in domains such as biomedicine, as in gene expression datasets~\citep{gao2015high} where there are 10K+ columns (different genes) and less than 100 rows (samples, e.g., patients). High-dimensionality of data has many solutions in the ARM literature, as it leads to rule explosion and, therefore, prolonged execution times. Existing methods include: i) mining rules for items of interest rather than all items, known as ARM with item constraints~\citep{srikant1997mining,baralis2012generalized,yin2022constraint}, ii) mining top-k high-quality rules based on a given rule quality criteria~\citep{fournier2012mining,nguyen2018etarm} and, iii) reducing rule redundancy by identifying only frequent itemsets without frequent supersets of equal support, known as closed itemset mining~\citep{zaki2002charm}. Aerial+~\citep{aerial_plus} (and the earlier version Aerial~\citep{aerial}), is a neurosymbolic method that is orthogonal to many of the existing solutions and leverages neural networks to learn a concise set of high-quality rules with full data coverage. Despite showing promising results on generic tabular datasets, it has not yet been evaluated on high-dimensional data. Furthermore, we argue that utilizing neural networks for ARM inherits neural networks-specific issues into ARM, most notably the reduced performance issue in low-data regimes~\citep{liu2017deep}. To the best of our knowledge, the low-data scenarios in ARM have not yet been addressed, as employing neural networks for ARM is a new paradigm shift. 

\begin{figure}[b]
    \centering
    \includegraphics[width=0.6\linewidth]{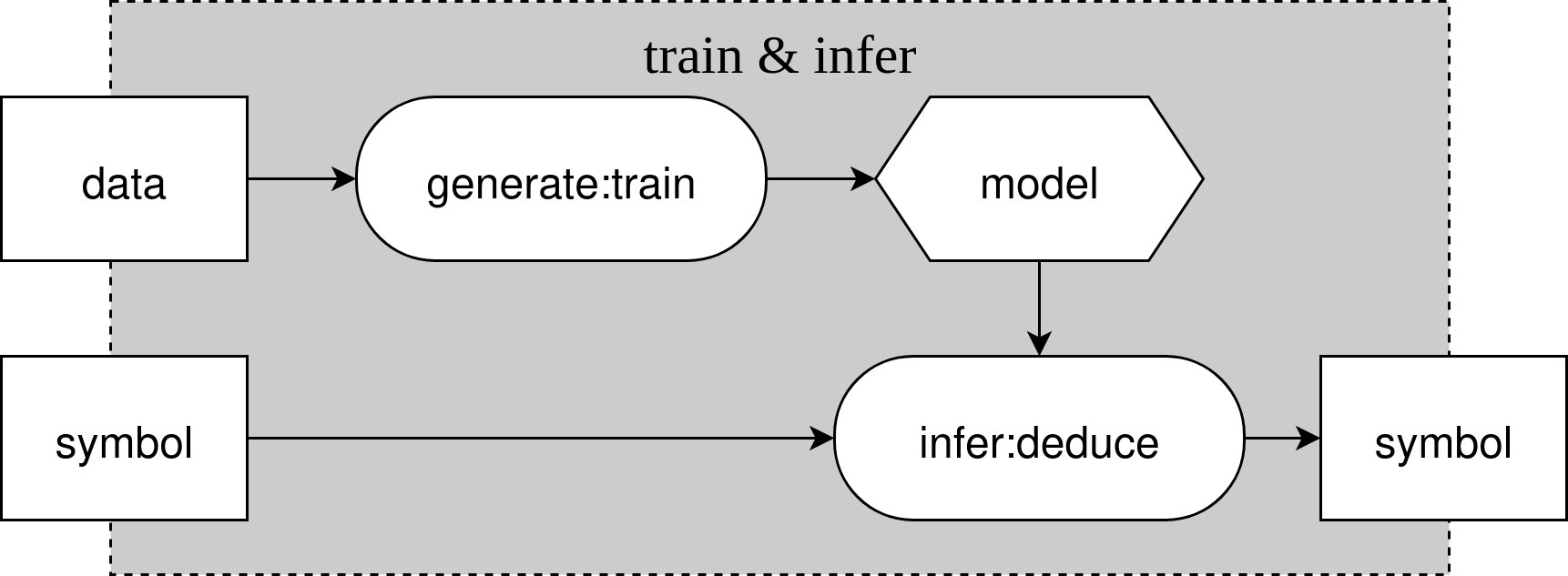}
    \vspace{5pt}
    \caption{Boxology~\citep{van2021modular} diagram of neurosymbolic ARM approaches such as Aerial+: i) a neural \textit{model} of \textit{data} (i.e., tabular data) is learned, ii) an algorithm (symbolic) \textit{infers} rules (symbols) from the model using hypotheses (symbols, as in test vectors of Aerial+).}
    \label{fig:nesy-arm-boxology}
\end{figure}

\textbf{Neurosymbolic methods for ARM.} Neural networks have been used to mine association rules directly from tabular data in the past few years. \citet{patel2022innovative} proposed the first approach that identifies frequent itemsets before constructing rules, but the work lacks an explicit algorithm or source code. \citet{berteloot2024association} introduced ARM-AE, an autoencoder-based~\citep{bank2023autoencoders} method to mine association rules directly. Aerial+~\citep{aerial_plus} tackles the rule explosion problem in ARM by using an under-complete denoising autoencoder~\citep{denoisingautoencoder} to learn a compact data representation, and by introducing a more scalable extraction method than ARM-AE (Figure~\ref{fig:aerial-pipeline}). This results in a smaller set of high-quality rules with full coverage over the data. Both Aerial+ and ARM-AE are neurosymbolic methods, combining neural models with symbolic rule extraction (Figure~\ref{fig:nesy-arm-boxology}). However, Aerial+ has not yet been evaluated on high-dimensional datasets, which we address in this work.

\textbf{Tabular foundation models} are large neural networks pre-trained on vast collections of tabular data to capture table semantics and support diverse downstream tasks~\citep{ruan2024language}. Among them, Tabular Prior-data Fitted Network (TabPFN)~\citep{tabpfn} is trained on millions of synthetic tables generated via structural causal models~\citep{pearl2009causality}, and supports classification and regression. Other recent tabular foundation models include CARTE~\citep{kim2024carte}, which leverages graph-based representations trained on real-world knowledge graphs~\citep{hogan2021knowledge}; TabICL~\citep{qu2025tabicl}, which frames tabular learning as in-context learning; Tabbie~\citep{iida2021tabbie}, which uses masked token modeling for pretraining; and TableGPT~\citep{su2024tablegpt2}, which adopts large language models for table understanding. Crucially, TabPFN is the only continuously maintained model that explicitly exposes an interface to extract table embeddings, which we utilized to develop fine-tuning strategies for Aerial+'s autoencoder architecture to learn higher-quality association rules in tables with a low number of rows~\footnote{We rely on the implementation available at \url{https://github.com/PriorLabs/TabPFN}.}. 

\section{ARM on high-dimensional datasets} \label{sec:arm-on-high-dim-data}

Given the focus on low-dimensional datasets in prior work on ARM, we begin with an empirical evaluation of the scalability of the state-of-the-art algorithmic and neurosymbolic ARM methods on high-dimensional categorical tabular datasets with few instances. Specifically, we aim to answer: \emph{how does the runtime cost of current ARM methods scale in the case of high-dimensional datasets?}

\textbf{Open-source.} All the source code and datasets used in all the experiments can be found in \url{https://github.com/DiTEC-project/rule_learning_high_dimensional_small_tabular_data}.

\textbf{Hardware.} All experiments are run on a 12th Gen Intel® Core™ i5-1240P × 16 CPU, with 16 GiB memory, and 512 GB disk space. No GPUs were used, and no parallel execution was conducted.

\textbf{Datasets.} We use 5 $d \gg n$ gene expression datasets from \citep{yang2012genomics,iorio2016landscape,garnett2012systematic,gao2015high} (listed in Table \ref{tab:dataset_sizes}), which are pre-processed according to the procedure described in \citep{mourragui2021predicting} by~\citep{ruiz2023high}. The pre-processing consists of the trimmed mean of m-values normalization, log transformation (i.e., $log(x+1)$), and the expression values were made to have zero mean and unit standard deviation. Furthermore, to enable ARM on gene expression datasets, we applied z-score binning with one standard deviation as the cutoff to discretize values into high, low, and medium gene expression levels, as exemplified in Table \ref{tab:motivation}.

\begin{table}[t]
    \centering
    \caption{High-dimensional tabular gene expression datasets with few instances, used in all experiments~\citep{yang2012genomics,iorio2016landscape,garnett2012systematic,gao2015high}.}
    \vspace{5pt}
    \begin{tabular}{lcc}
        \toprule
        \textbf{Dataset} & \textbf{\# Columns} & \textbf{\# Rows} \\
        \midrule
        Chondrosarcoma & 18006 & 6 \\
        SmallCellLungCarcinoma & 18237 & 60 \\
        NonSmallCellLungCarcinoma & 18108 & 86 \\
        BreastCarcinoma & 18061 & 51 \\
        Melanoma & 17902 & 55 \\
        \bottomrule
    \end{tabular}
    \label{tab:dataset_sizes}
\end{table}

\textbf{Algorithms.} We run the state-of-the-art neurosymbolic ARM method Aerial+~\citep{aerial_plus}, commonly used algorithmic methods, ECLAT~\citep{eclat} and FP-Growth~\citep{han2000mining}, as well as ARM-AE~\citep{berteloot2024association} on all the datasets given in Table \ref{tab:dataset_sizes}. FP-Growth remains one of the most widely used ARM algorithms due to its efficiency and adaptability. Numerous variations to FP-Growth have been proposed to mitigate rule explosion and improve scalability, including Guided FP-Growth~\citep{guidedfpgrowth} for item-constrained mining, parallel FP-Growth~\citep{parallelfpgrowth}, and GPU-accelerated versions~\citep{gpufpgrowth} for faster execution. Note that Aerial+ also supports item constraints, parallel, and GPU executions. However, we only compare the basic version of each algorithm. 

\begin{table}[b]
    \centering
    \caption{Evaluated ARM algorithms and hyperparameters for fair comparison on tabular data (R = Aerial+ rules, C = Columns).}
    \vspace{5pt}
    \label{tab:algorithms}
    \begin{tabular}{lcc}
        \toprule
        \textbf{Algorithm} & \textbf{Type} & \textbf{Parameters} \\
        \midrule
        Aerial+ & Neurosymbolic & $a=2, \tau_a=0.5, \tau_c=0.8$ \\
        ARM-AE & Neurosymbolic & M=2, N=$|R|/|C|$, L=0.5 \\
        \midrule
        FP-Growth & Algorithmic & \multirow{2}{0.56\linewidth}{antecedents = 2, min\_conf=0.8, min\_support=$0.5 \ast \mathbb{E}[\text{support}(R)]$} \\
        ECLAT & Algorithmic & \\
        \bottomrule
    \end{tabular}
\end{table}

\textbf{Experimental setup and hyperparameters.} To ensure a fair comparison, we set the hyperparameters of each method (shown in Table \ref{tab:algorithms}) as follows: i) number of antecedents is set to 2 for all methods, ii) Aerial+'s antecedent similarity threshold ($\tau_a$) and ARM-AE's likeness ($L$) are set to 0.5, iii) Aerial+'s consequent similarity threshold ($\tau_c$) and minimum confidence of the algorithmic methods are set to 0.8, iv) minimum support threshold of the algorithmic methods are set to half the average support of the rules learned by Aerial+, to ensure comparable average support values, v) ARM-AE's number of rules per consequent ($N$) is set to Aerial+'s rule count divided by the number of columns to ensure comparable rule counts, vi) and both Aerial+ and ARM-AE were trained for 10 epochs with a batch size of 2. Aerial+ is implemented using the pyAerial~\footnote{https://github.com/DiTEC-project/pyaerial}~\citep{pyaerial} library, FP-Growth is implemented using MLxtend~\citep{raschkas_2018_mlxtend}, ECLAT is implemented using pyECLAT~\footnote{https://github.com/jeffrichardchemistry/pyECLAT}, and ARM-AE is implemented using its original repository~\footnote{https://github.com/TheophileBERTELOOT/ARM-AE/tree/master}. The \textbf{goal} of this experimental setup is to test the scalability of the algorithms, and not to perform a rule quality comparison, which has already been done in earlier work~\citep{aerial_plus}.

\begin{figure*}[t]
    \centering
    \includegraphics[width=\linewidth]{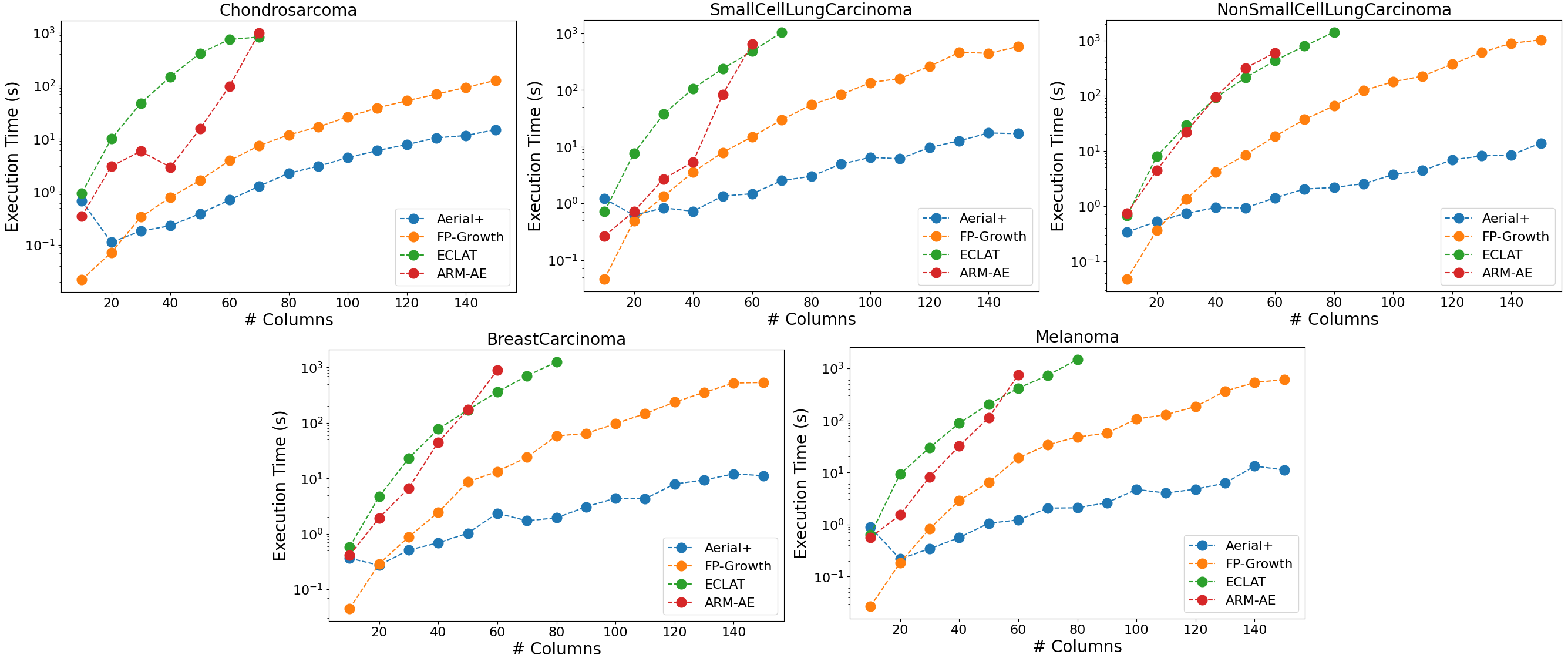}
    \caption{\justifying\textbf{Scalability on high-dimensional tabular data.} Execution times of algorithmic and neurosymbolic (including training and rule extraction time) ARM approaches in seconds on a logarithmic scale, as the number of columns increases gradually. Aerial+ has one to two orders of magnitude better scalability on high-dimensional datasets compared to other methods. Lower performance of Aerial+ with a smaller number of columns is due to the training procedure, which implies that algorithmic methods are faster on lower-dimensional (columns) tables.}
    \label{fig:exec-time}
\end{figure*}

\textbf{Results.} Figure \ref{fig:exec-time} shows the execution time of each method, in seconds on a logarithmic scale, on 5 datasets as the number of columns increases. Execution times include both training and the rule extraction times for the neurosymbolic methods. The results show that Aerial+ has one to two orders of magnitude faster execution times than the other methods. The gap in execution time increases as the number of columns increases. We also see that the algorithmic method FP-Growth runs faster when the number of columns is smaller than 30. This shows that Aerial+'s training is only compensated if the tables have more than 30 columns. Note that Aerial+ has linear time complexity during training and polynomial time over the number of columns (after one-hot encoding) during the extraction.

\section{Neurosymbolic ARM in low-data regime} \label{sec:arm-in-low-data}

Experiments in Section \ref{sec:arm-on-high-dim-data} showed that the fastest algorithmic solution, FP-Growth, takes \textasciitilde$10^3$ seconds on tables with only 150 columns and 2 antecedents, while a neurosymbolic method, Aerial+, runs one to two orders of magnitude faster. This empirically validates the scalability of neurosymbolic approaches to ARM. However, we argue that Aerial+ also inherits the known issues in neural networks, particularly the decline in performance in a low-data regime~\citep{liu2017deep}. Concretely, Aerial+ relies on training a deep autoencoder on the tabular data with a reconstruction objective. Following results from statistical learning theory~\citep{vapnik1998statistical} and empirical observations in neural networks~\citep{zhang2016understanding}, this implies that Aerial+’s performance is bounded by the number of training samples, and with small data it may yield rules that do not accurately capture ground-truth associations.

An effective approach for addressing data scarcity is \emph{transfer learning}~\citep{weiss2016transfer}, which requires training a neural network, or vector representations (i.e. \emph{embeddings}) on a large dataset, that then can be transferred to a downstream task on a small dataset. This provides a starting point that can improve performance in comparison to learning from scratch on a small dataset. 

In this work, we propose two fine-tuning strategies to Aerial+ using TabPFN~\citep{tabpfn}, a foundation model for tabular data that has been pre-trained over millions of tables, which we use to generate embeddings for the small datasets in our experiments.

\subsection{Fine Tuning with Pre-trained Weight Initialization} \label{sec:ft-weight-init}

\begin{figure}[t]
    \centering
    \includegraphics[width=\textwidth]{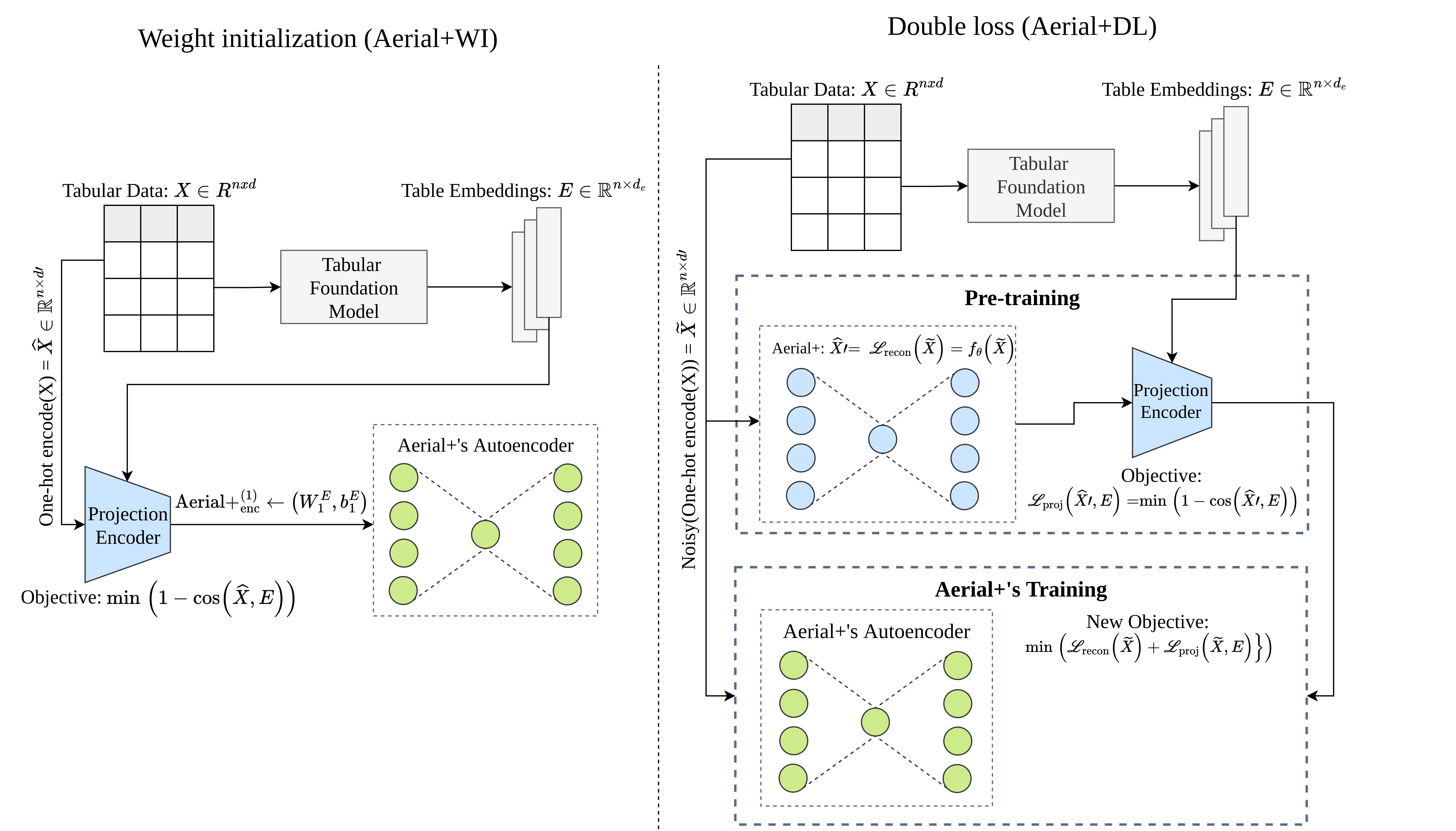}
    \caption{\justifying\textbf{Weight initialization (Aerial+WI, Left)}: tabular data is embedded using a foundation model, then a projection encoder is trained to align these embeddings with pre-processed Aerial+ input. The learned projection encoder is used to initialize the first-layer weights and biases of Aerial+'s encoder, providing a semantically meaningful starting point for fine-tuning.     \textbf{Double loss (Aerial+DL, Right)}: tabular embeddings are aligned with reconstructed Aerial+ outputs using a projection encoder, and this alignment objective is incorporated into the Aerial+ autoencoder reconstruction loss. This double loss encourages the autoencoder to produce reconstructions semantically consistent with the original table embeddings, supporting accurate fine-tuning.}
    \label{fig:aerial-finetuned}
\end{figure}

Figure~\ref{fig:aerial-finetuned} (left) illustrates the fine-tuning strategy introduced in this section (Aerial+WI). On a high level, table embeddings from a tabular foundation model are utilized to initialize the weights of Aerial+’s under-complete denoising autoencoder, providing a semantically meaningful starting point for learning compact data representations.

Let $X \in \mathbb{R}^{n \times d}$ denote the tabular dataset and $y \in \mathbb{R}^n$ the corresponding labels. We first compute fixed-length embeddings for each row in $X$ using a pretrained TabPFNClassifier. These embeddings, denoted as $E \in \mathbb{R}^{n \times d_e}$, where $d_e$ is the embedding dimension, are generated via a 10-fold TabPFN-based meta-learning scheme:

\[
E = f_{\text{TabPFNClassifier}}(X, y)
\]

We then one-hot encode $X$ into $\hat{X} \in \mathbb{R}^{n \times d'}$ following the original Aerial+ pipeline, where $d'$ is the total number of binary features after encoding categorical attributes. A two-layer projection encoder $g_\theta: \mathbb{R}^{d'} \rightarrow \mathbb{R}^{d_e}$ is trained to map $\hat{X}$ to the TabPFN embedding space. The encoder architecture is as follows:
\[
g_\theta(\hat{x}) = W_2 \cdot \text{Dropout}(\sigma(\text{LayerNorm}(W_1 \hat{x} + b_1))) + b_2
\]
where $W_1 \in \mathbb{R}^{h \times d'}$, $W_2 \in \mathbb{R}^{d_e \times h}$, $h$ is the hidden dimension, $\sigma$ is the LeakyReLU activation with a negative slope of 0.01, and LayerNorm and Dropout ($p=0.1$) are applied for regularization.

The projection encoder is trained using a cosine loss function to align $g_\theta(\hat{X})$ with $E$:

\[
\mathcal{L}(\theta) = 1 - \frac{1}{n} \sum_{i=1}^n \cos\left(g_\theta(\hat{x}_i), E_i\right)
\]
Training is performed using Adam optimizer~\citep{kingma2014adam} for 25 epochs with early stopping if the validation loss plateaus (with early stopping patience of 20 and a minimum improvement threshold of $10^{-4}$). After training, the weight matrix $W_1$ and bias $b_1$ from the first layer of $g_\theta$ are used to initialize the corresponding parameters in the first layer of Aerial+'s encoder:

\[
\text{Aerial+}_\text{enc}^{(1)} \gets (W_1, b_1)
\]

This initialization provides a strong inductive prior for Aerial+, guiding its encoder to start from a semantically meaningful representation space derived from TabPFN's meta-learned embeddings.

Note that the gene expression datasets contain no predefined class labels. Therefore, a random column is selected as the target variable to enable TabPFN embedding generation.

\subsection{Projection-Guided Fine Tuning via Double Loss}
\label{subsec:double-loss}

Figure \ref{fig:aerial-finetuned} (right) visualizes the fine-tuning strategy described in this section (Aerial+DL). Conceptually, this strategy uses a projection encoder to align Aerial+ reconstructions with table embeddings from a tabular foundation model, jointly optimizing reconstruction and alignment losses for semantic consistency.

Building on the projection encoder $g_\theta$ described in Section \ref{sec:ft-weight-init}, this second fine-tuning strategy aligns the Aerial+'s autoencoder reconstructions with TabPFN embeddings using a double loss function.

Unlike the first strategy, where $g_\theta$ was trained directly on raw one-hot inputs, here we first pass a corrupted version of the one-hot input $\hat{X}$ through Aerial+'s initial autoencoder $f_\theta$ and train $g_\theta$ on its outputs. Specifically, we generate noisy inputs (following the same strategy as Aerial+):
\[
\tilde{X} = \text{clip}(\hat{X} + \epsilon), \quad \epsilon \sim \mathcal{N}(0, \sigma^2)
\]
where $\sigma = 0.5$ and values are clipped to $[0,1]$. We then compute reconstructions $\hat{X}' = f_\theta(\tilde{X})$. The projection encoder is trained to map these reconstructions to their corresponding TabPFN embeddings $E \in \mathbb{R}^{n \times d_e}$:
\[
g_\theta(\hat{x}'_i) \approx E_i
\]
by minimizing the cosine distance loss:
\[
\mathcal{L}_{\text{proj}}(\theta) = 1 - \frac{1}{n} \sum_{i=1}^n \cos(g_\theta(\hat{x}'_i), E_i)
\]

After this pretraining phase, $g_\theta$ is frozen, and Aerial+'s autoencoder is fine-tuned using a \emph{double loss} objective:
\[
\mathcal{L}(\theta) = \mathcal{L}_{\text{recon}}(f_\theta(\tilde{x}), \hat{x}) + \mathcal{L}_{\text{proj}}(g_\theta(f_\theta(\tilde{x})), E)
\]
where $\mathcal{L}_{\text{recon}}$ is a binary cross-entropy loss applied to per one-hot encoded column value as in Aerial+, generating probability distributions per column. The double loss strategy encourages Aerial+'s autoencoder to not only reconstruct the original data, but also to produce representations that are semantically consistent with TabPFN's meta-learned embedding space.

\subsection{Experimental Results}

\begin{table*}[t]
    \centering
    \setlength{\tabcolsep}{2pt}
    \caption{\justifying\textbf{Rule quality of Aerial+ in low-data regime.} Fine-tuning Aerial+ with the weight initialization (Aerial+WI) and double loss (Aerial+DL) methods based on TabPFN embeddings consistently outperformed the default version in rule confidence and association strength (Zhang's). Fine-tuning produced fewer rules with lower data coverage on 3 of 5 datasets, as expected due to the elimination of relatively obvious (low-association-strength) rules. Execution time increased by only a few seconds, which is negligible in a low-data regime.}
    \begin{tabular}{lccccccc}
        \toprule
        \textbf{Approach} & \textbf{\# Rules} & \textbf{\textasciitilde Rule Coverage} & \textbf{\textasciitilde Support} & \textbf{\textasciitilde Confidence} & \textbf{Data Coverage} & \textbf{\textasciitilde Zhang's Metric} & \textbf{Exec. Time (s)} \\
        \midrule
        \multicolumn{8}{c}{\textbf{Chondrosarcoma}} \\
        Aerial+ & 200 & \textbf{0.23} & 0.21 & 0.921 & 0.533 & 0.784 & 2.25 \\
        Aerial+WI & 75 & 0.217 & 0.206 & 0.945 & 0.524 & 0.813 & 5.80 \\
        Aerial+DL & 75 & 0.235 & 0.219 & \textbf{0.947} & \textbf{0.536} & \textbf{0.828} & 5.36 \\
        \midrule
        \multicolumn{8}{c}{\textbf{SmallCellLungCarcinoma}} \\
        Aerial+ & 1576 & 0.068 & 0.041 & 0.579 & \textbf{0.835} & 0.476 & 10.58 \\
        Aerial+WI & 664 & \textbf{0.076} & 0.052 & \textbf{0.633} & 0.715 & \textbf{0.577} & 13.48 \\
        Aerial+DL & 1338 & 0.070 & 0.044 & 0.597 & 0.816 & 0.513 & 18.23 \\
        \midrule
        \multicolumn{8}{c}{\textbf{NonSmallCellLungCarcinoma}} \\
        Aerial+ & 1620 & 0.059 & 0.035 & 0.584 & 0.823 & 0.554 & 18.03 \\
        Aerial+WI & 978 & \textbf{0.078} & 0.057 & \textbf{0.663} & 0.698 & \textbf{0.639} & 28.67 \\
        Aerial+DL & 1453 & 0.053 & 0.028 & 0.547 & \textbf{0.849} & 0.501 & 24.27 \\
        \midrule
        \multicolumn{8}{c}{\textbf{BreastCarcinoma}} \\
        Aerial+ & 1017 & 0.072 & 0.046 & 0.641 & \textbf{0.816} & 0.575 & 9.64 \\
        Aerial+WI & 590 & 0.077 & 0.052 & \textbf{0.686} & 0.686 & \textbf{0.644} & 12.09 \\
        Aerial+DL & 535 & \textbf{0.078} & 0.050 & 0.652 & 0.761 & 0.590 & 15.31 \\
        \midrule
        \multicolumn{8}{c}{\textbf{Melanoma}} \\
        Aerial+ & 1220 & 0.067 & 0.035 & 0.545 & \textbf{0.888} & 0.440 & 13.09 \\
        Aerial+WI & 773 & 0.070 & 0.038 & \textbf{0.575} & 0.772 & \textbf{0.496} & 13.19 \\
        Aerial+DL & 859 & \textbf{0.071} & 0.038 & 0.566 & 0.860 & 0.461 & 16.49 \\
        \bottomrule
    \end{tabular}
    \label{tab:arm_results_summary}
\end{table*}

\textbf{Setup and hyperparameters.} We run Aerial+ and the two fine-tuned versions, with pre-trained weight initialization (Aerial+WI) and double loss (Aerial+DL), on 5 $d \gg n$ datasets with 100 columns and compare their rule quality. The default Aerial+ uses Xavier~\citep{glorot2010understanding} weight initialization as in the original work. All the approaches are run with 2 antecedents, for 25 epochs with a batch size of 2. Aerial+'s autoencoder for both the default and the fine-tuned versions consists of 2 layers per encoder and decoder, with the dimensions $\hat{X} \rightarrow 50 \rightarrow 10$, and the mirrored version for the decoder. We run each method \textbf{50 times} and present the average rule quality results for robustness.

\textbf{Evaluation criteria.} The standard rule quality metrics from the ARM literature are used as the evaluation criteria~\citep{luna2019ARMsurvey,kaushik2023numerical}. Let $D$ be a set of transactions as introduced in Section \ref{sec:related-work}, and $R = \{r_1, r_2, ..., r_t\}$ be the rule set learned by each approach where $\forall r_i \in R, r_i = (X_i \rightarrow Y_i)$: 
\begin{itemize}
    \item \textbf{Number of rules.} Total number of rules learned:$|R|$
    \item \textbf{Average rule coverage.} Average number of transactions where the rule antecedent appears: $\text{AvgCov} = \frac{1}{|R|} \sum_{i=1}^{|R|} |\{ t \in D \mid X_i \subseteq t \}|$

    \item \textbf{Average support.} Average fraction of transactions containing both antecedent and consequent:
    \[
    \text{AvgSupp} = \frac{1}{|R|} \sum_{i=1}^{|R|} \frac{|\{ t \in D \mid X_i \cup Y_i \subseteq t \}|}{|D|}
    \]

    \item \textbf{Average confidence.} Average conditional probability that the consequent appears given the antecedent:
    \[
    \text{AvgConf} = \frac{1}{|R|} \sum_{i=1}^{|R|} \frac{|\{ t \in D \mid X_i \cup Y_i \subseteq t \}|}{|\{ t \in D \mid X_i \subseteq t \}|}
    \]

    \item \textbf{Total data coverage.} Fraction of transactions covered by at least one rule antecedent:
    \[
    \text{TotalCov} = \frac{\left| \bigcup_{i=1}^{|R|} \{ t \in D \mid X_i \subseteq t \} \right|}{|D|}
    \]

    \item \textbf{Average Zhang's metric~\citep{zhangsmetric}.} Average statistical dependence between antecedent and consequent beyond chance:
    \[
    \text{AvgZhang} = \frac{1}{|R|} \sum_{i=1}^{|R|} \text{Zhang}(X_i \Rightarrow Y_i)
    \]
    where:
    \[
        \text{Zhang}(X_i \rightarrow Y_i) = \frac{\text{conf}(X_i \rightarrow Y_i) - \text{conf}(X_i^{'} \rightarrow Y_i)}{max(\text{conf}(X_i \rightarrow Y_i), \text{conf}(X_i^{'} \rightarrow Y_i))}
    \]
    with conf being the confidence score of a rule and $X_i^{'}$ referring to the absence of $X_i$ in $D$.
    \item \textbf{Execution time.} Sum of model training time, fine-tuning (when applicable), and rule extraction time in seconds. 
\end{itemize}

\textbf{Results.} Table \ref{tab:arm_results_summary} shows the rule quality evaluation results of Aerial+ and the two fined-tuned versions Aerial+WI and Aerial+DL on 5 datasets. The results show that Aerial+WI outperforms Aerial+ in terms of rule confidence and association strength (Zhang's metric) on all 5 datasets. Aerial+DL's confidence and association strength also exceed Aerial+'s on 4 out of 5 datasets, except the NonSmallCellLungCarcinoma dataset. Both fine-tuning methods resulted in a smaller number of rules on all datasets and with a smaller data coverage on 3 out of the 5. This is expected as the fine-tuned versions capture rules with higher association strength on average, meaning the less obvious rules are eliminated during the rule extraction process, and therefore, the final data coverage was lower. The fine-tuned methods have higher support values on 4 out of 5 datasets. However, we do not take the high support values as a positive sign, as it depends on the application. For instance, high support rules are good at explaining trends in the data, while low support rules can be better at explaining anomalies. Lastly, fine-tuning resulted in only a few seconds of increment in the execution time, which is negligible in the low-data regime. Note that the costliest operation in Aerial+ is the rule extraction process and not the training (or pre-training), which is not significantly affected by the fine-tuning methods.

\section{Discussion} \label{sec:discussion}

\vspace{-5pt}

The section discusses the experimental results, the role of neurosymbolic and tabular foundation models in ARM.

\textbf{Neurosymbolic methods scale better on high-dimensional data.} Experiments in Section \ref{sec:arm-on-high-dim-data} show that Aerial+, a neurosymbolic method to ARM, has execution speed of one to two orders of magnitude faster than the algorithmic ARM approaches. We argue this is because Aerial+ leverages neural networks' ability to handle high-dimensional data, it has linear complexity over the number of rows in training, and polynomial time complexity over the number (one-hot encoded) columns during the rule extraction stage. Algorithmic methods, on the other hand, rely on counting the co-occurrences of itemsets in the data, which is a costlier operation.

\textbf{Aerial+ inherits neural networks-specific issues into ARM.} The scalability of Aerial+ on high-dimensional data comes at a cost, most notably the reduced performance in the low-data regime for ARM. The original paper of Aerial+ trains only for 2 epochs on generic tabular datasets and was able to obtain high-quality rules. In the low-data regime, however, we were able to get high-quality rules consistently in each execution only after training for 25 epochs. This shows that while the neurosymbolic methods can help in scalability, they also introduce a new research problem into the ARM literature, namely, rule mining in the low-data regime.

\textbf{Fine-tuning Aerial+ for better knowledge discovery.} Experiments in Section \ref{sec:arm-in-low-data} showed that our two proposed fine-tuning methods using the tabular foundation model TabPFN resulted in significantly higher-quality rules in comparison to the default version of Aerial+ on 5 real-world high-dimensional tabular datasets with few instances. Many of the other tabular foundation models that we investigated, including Tabbie, CARTE, TableGPT, and TabICL, do not provide an interface to obtain table embeddings. Therefore, we were not able to use them in our experiments. Since TabPFN is trained to perform classification and regression tasks over tabular data, we expect that models explicitly trained to learn column embeddings and associations could potentially result in better rule quality. 

\textbf{Neurosymbolic methods start a paradigm shift in ARM.} We show that the Neurosymbolic ARM methods can be supported by prior-data fitted networks, as in TabPFN, to learn higher-quality rules. This raises the research question of \textbf{what other types of prior data or background knowledge can be utilized as part of ARM?} We invite researchers to further investigate neurosymbolic methods for ARM, as the neurosymbolic integration brings an immense potential for both knowledge discovery and fully interpretable inference across a plethora of domains. 

\textbf{Further validation of our approach and limitations.} The algorithmic methods strictly depend on the distribution of data when mining rules in terms of execution time, as \textit{denser} datasets, where many frequent itemsets of high support are present, will eventually prolong the execution time. Aerial+, however, applies the exact same polynomial-time rule extraction process regardless of the density of the data, and therefore depends less on the dataset attributes. However, we will still test our fine-tuning approaches on more datasets from diverse domains to further validate our approach in future work. Furthermore, we will evaluate our approach on generic tabular data with higher numbers of instances, i.e., $n \gg d$, to see whether it leads to early convergence or higher quality rules. Our proposed fine-tuning strategies are currently limited to the only available tabular foundation model with an explicit table embedding interface, TabPFN. Since TabPFN is specifically trained for classification and regression, this limitation may restrict performance improvements, and a future foundation model trained to capture column associations explicitly could significantly improve rule discovery.

\section{Conclusions}

\vspace{-5pt}

This paper highlights the potential of neurosymbolic methods in the domain of association rule mining (ARM), especially under high-dimensional and low-sample ($d \gg n$) settings common in domains such as biomedicine. We have empirically shown that Aerial+, a neurosymbolic approach, offers substantial scalability improvements compared to the state-of-the-art neurosymbolic and algorithmic ARM techniques, scaling one to two orders of magnitude faster. However, neurosymbolic ARM also inherits the known issues of neural networks into ARM literature, specifically the reduced performance in low-data regimes, which we addressed through two targeted fine-tuning strategies.

Our fine-tuning methods use table embeddings from TabPFN, a tabular foundation model, to i) initialize the weights of Aerial+ (Aerial+WI), ii) and to better semantically align Aerial+ autoencoder training with a given tabular data (Aerial+DL). The results show that both Aerial+WI and Aerial+DL methods significantly improved rule quality in low-data settings. This demonstrates the promising role of pretrained tabular models in enhancing knowledge discovery over tabular datasets, besides classification and regression tasks that are commonly tackled in the tabular data domain. 

Looking forward, we see this as the beginning of a broader paradigm shift in ARM, where background knowledge and pretrained models can be explicitly leveraged to guide rule extraction. We invite the community to explore what other forms of prior knowledge, architectures, or foundation models can be integrated into neurosymbolic ARM. Future work will also validate our methods across a wider range of datasets and evaluate their effectiveness in high-instance scenarios ($n \gg d$), with the aim of achieving both scalability and high interpretability in real-world data mining applications.

\section*{Acknowledgements}
\label{sec:acknowledgements}
This work has received support from the Dutch Research Council (NWO), in the scope of the Digital Twin for Evolutionary Changes in water networks (DiTEC) project, file number 19454.

\bibliographystyle{unsrtnat}
\bibliography{main}

\appendix

\end{document}